\documentclass[letterpaper, 10 pt, conference]{ieeeconf}
\IEEEoverridecommandlockouts 
\usepackage{times}

\usepackage{multicol}
\usepackage[bookmarks=true]{hyperref}

\let\labelindent\relax

\usepackage{amsmath,amssymb,amsthm}
\usepackage{algpseudocode,algorithm}
\usepackage{mathtools}
\usepackage{enumitem}

\newtheorem{theorem}{Theorem}[section]
\newtheorem{definition}[theorem]{Definition}
\newtheorem{assumption}[theorem]{Assumption}
\newtheorem{lemma}[theorem]{Lemma}
\newtheorem{remark}[theorem]{Remark}

\newcommand{\X}{\mathcal{X}}

\newcommand{\U}{\mathcal{U}}

\newcommand{\Xs}{\X_{\text{safe}}}
\newcommand{\Xinv}{\X_{\text{inv}}}
\newcommand{\Xr}{\X_{\text{rec}}}

\newcommand{\Xe}{\X_{\text{eq}}}

\usepackage{color}

\title{\LARGE \bf Safe Human-Interactive Control via Shielding}

\author{Jeevana Priya Inala$^{1}$, Yecheng Jason Ma$^{2}$, Osbert Bastani$^{2}$, Xin Zhang$^{3}$, Armando Solar-Lezama$^{1}$
\thanks{$^{1}$Jeevana Priya Inala and Armando Solar-Lezama are with the Department of Electrical Engineering and Computer Science, Massachusetts Institute of Technology, Cambridge, MA 02139, USA {\tt\small \{jinala, asolar\}@csail.mit.edu}}%
\thanks{$^{2}$Yecheng Jason Ma and Osbert Bastani are with the Department of Computer and Information Science, University of Pennsylvania, Philadelphia, PA 19104, USA {\tt\small \{jasonyma,obastani\}@seas.upenn.edu}}%
\thanks{$^{3}$Xin Zhang is with the Department of Computer Science and Technology at Peking University, Beijing, 100871, China {\tt\small xin@pku.edu.cn}}%
}
\begin{document}

\maketitle
\thispagestyle{empty}
\pagestyle{empty}

\begin{abstract}
Ensuring safety for human-interactive robotics is  important due to the potential for human injury. The key challenge is defining safety in a way that accounts for the complex range of human behaviors without modeling the human as an unconstrained adversary. We propose a novel approach to ensuring safety in these settings. Our approach focuses on defining \emph{backup actions} that we believe human always considers taking to avoid an accident---e.g., brake to avoid rear-ending the other agent. Given such a definition, we consider a safety constraint that guarantees safety as long as the human takes the appropriate backup actions when necessary to ensure safety. Then, we propose an algorithm that overrides an arbitrary given controller as needed to ensure that the robot is safe. We evaluate our approach in a simulated environment, interacting with both real and simulated humans.\footnote{Our full paper (with appendix): \url{https://https://obastani.github.io/icra22.pdf}.}

\end{abstract}


\section{Introduction}
\label{sec:intro}

Robots are increasingly operating in environments where they must interact with humans, such as collaborative grasping~\cite{strabala2013toward,dragan2013legibility} and autonomous driving~\cite{montemerlo2008junior,levinson2011towards,sadigh2016planning,sadigh2016information}. Thus, there has been much interest in designing planning and control algorithms for human-robot interaction. Ensuring safety for such robots is paramount due to the potential to inflict harm on humans~\cite{eder2014towards}. These challenges are particularly salient in settings such as autonomous driving, where robots and humans may have disjoint or conflicting goals---e.g., a self-driving car making an unprotected left turn~\cite{sadigh2016planning}.

The key challenge is how to define safety for human-interactive robots. We could model the human as an adversary, but this approach is typically prohibitively conservative. Another approach is to learn a model to predict human actions~\cite{ziebart2008maximum,fisac2018probabilistically,sadigh2019verifying}, and ensure safety with respect to this model. If the model captures all actions exhibited by humans, then this approach ensures safety. However, different humans may exhibit very different behaviors~\cite{sadigh2016information}---e.g., people in a lab may act differently than people on a street. Collecting data from all possible settings can be very challenging. If a behavior is not exhibited in the training data, then the model may not account for it. More fundamentally, even the best machine learning models make errors, which can correspond to actions missed by the model. Finally, \emph{responsibility sensitive safety (RSS)}~\cite{shalev2017formal} is an approach where that manually specifies the range of acceptable robot actions in various scenarios. That is, the designer of the robot controller is responsible for ensuring that acceptable actions only include safe actions. However, manually defining acceptable robot actions for all possible scenarios is challenging, especially for robots operating in open-world environments. For instance,~\cite{shalev2017formal} only formally defines acceptable actions for a limited number of scenarios such as changing lanes.

\begin{figure}
\centering
\includegraphics[height=0.22\textwidth]{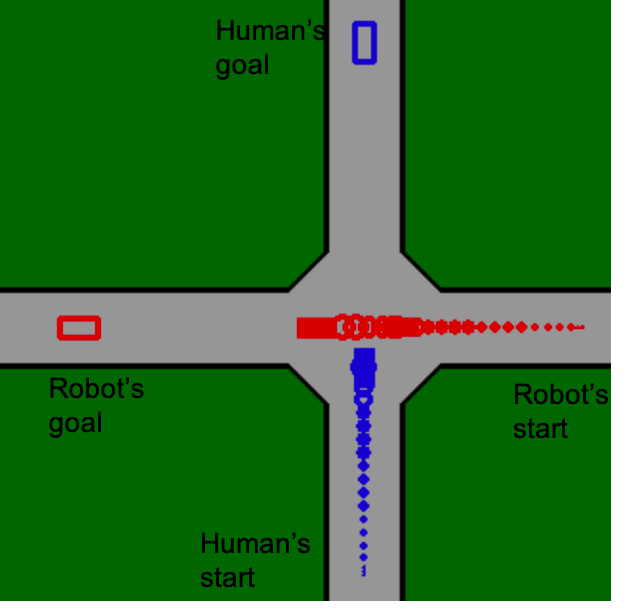}
\includegraphics[height=0.22\textwidth]{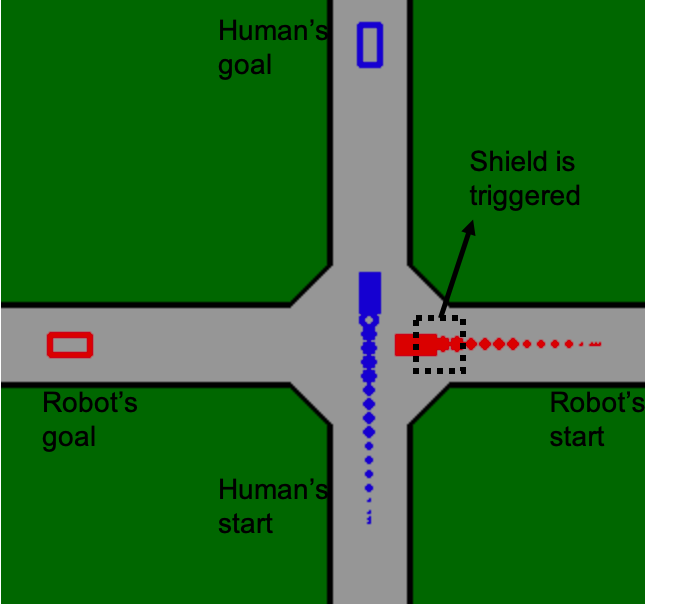}
\caption{Trajectories showing a robot (red) and a human (blue) interacting at an intersection (for 25 timesteps). Left: The robot passes before the human, leveraging the fact that a responsible human would slightly brake to allow the robot to cross safely. Right: Human arrives at the intersection first; the robot triggers the shield to brake and  allow the human to cross first.}
\label{fig:overview}
\end{figure}

We propose a novel approach for ensuring safety in human-interactive systems that accounts for all human behaviors in some bounded set. There are two challenges: (i) how to define the set of human behaviors, and (ii) how to ensure safety with respect to this set. We address these challenges as follows:
\begin{itemize}
\item {\bf Bounding human behavior via backup actions:} Rather than specify the set of all actions the robot is \emph{allowed} to take (as in RSS), the designer specifies \emph{backup actions} that we believe the human always considers taking to avoid an accident (e.g., braking while steering in some direction). In particular, we assume the human may take any action in general, but that they take these actions when necessary to ensure safety.
\item {\bf Ensuring safety via abstract interpretation:} We use \emph{abstract interpretation}~\cite{cousot1977abstract} to conservatively overapproximate the reachable set of the system for the above model of human behavior, and then ensure safety with respect to this overapproximation.
\end{itemize}
First, our notion of backup actions captures the idea that we reasonably believe the human will take a limited range of evasive maneuvers to avoid an accident---e.g., if the robot gradually slows to a stop, then we may expect the human to also slow down to avoid rear-ending it.\footnote{We do not assume the human will take evasive maneuvers such as swerving to avoid an accident (though can do so if it is safe). Also, we assume a reasonable amount of time for them to brake and come to a stop.}
If the robot is on a highway, coming to a stop is more dangerous, so we may require that the robot pull over to the shoulder before coming to a stop. Similarly, we may require that the robot avoid coming to a stop in an intersection. Specifying backup actions provides a way to define safety; we refer to such a safety constraint as \emph{safety modulo fault}.\footnote{The term ``fault'' is used simply to mean that safety ignores cases where the human does not act according to our model, not to blame an accident on the human driver. The controller designer is responsible for designing the backup actions to obtain a reasonable notion of safety.}

At a high level, to instantiate our framework, the designer of the robot controller needs to design the following:
\begin{itemize}
\item \textbf{Robot backup action:} An action the human anticipates the robot may take to ensure safety---e.g., brake without changing directions. It should be chosen based on intuition about what actions the human driver anticipates the robot may take (e.g., based on traffic rules).
\item \textbf{Human backup action set:} A set of actions that includes \emph{at least one} action the human considers taking to ensure safety---e.g., braking while steering in some direction. It should be chosen based on intuition about safety maneuvers the human driver considers taking.
\end{itemize}
In particular, whereas RSS requires the control designer to specify the set of \emph{all} actions the robot is allowed to take, we only require the designer to specify actions that they believe the human considers taking to use to avoid an accident.

Next, we propose an algorithm for ensuring safety modulo fault. 
Our algorithm builds on \emph{shielding}~\cite{akametalu2014reachability,alshiekh2018safe,wabersich2018linear,li2019robust}; in particular, rather than designing a specific controller, our algorithm takes as input an arbitrary controller designed to achieve the goal, and then combines it with backup controller in a way that tries to use the given controller as frequently as possible while ensuring safety modulo fault. At a high level, it does so by using on-the-fly verification based on abstract interpretation to determine whether it is safe modulo fault to use the given controller; if so, it uses the given controller, but otherwise, it uses a backup controller.

Finally, we empirically evaluate our approach in a simulation, including both settings where the humans are simulated and settings where the human is controlled by an real person via keyboard inputs. We demonstrate that our MPS modulo fault algorithm enables the robot to avoid accidents both with real and simulated humans, even when combined with a na\"{i}ve controller that altogether ignores the humans.

Figure~\ref{fig:overview} illustrates how our algorithm ensures safety while interacting with a human driver without being overly cautious. It assumes that the human will at least slightly brake to avoid an accident (left). If it still cannot guarantee safety, then it allows the human to go first (right).

\section{Preliminaries}

\paragraph{Human-robot system}

We consider a system with a robot $R$ and human $H$. Following prior work~\cite{sadigh2016planning}, we assume the human and robot act in alternation, which is reasonable as long as the time steps are sufficiently small. For a state $x_t$ where it is the robot's turn to act, we have 
\begin{align*}
x_t'=f_R(x_t,u_{R,t})\qquad\text{and}\qquad x_{t+1}=f_H(x_t',u_{H,t}),
\end{align*}
where $f_R:\X\times\U_R\to\X$ is the robot dynamics, $f_H:\X\times\U_H\to\X$ is the human dynamics, $\X\subseteq\mathbb{R}^{n_X}$ is the joint state space, $\U_R\subseteq\mathbb{R}^{n_{U,R}}$ are the robot actions, and $\U_H\subseteq\mathbb{R}^{n_{U,H}}$ are the human actions. Given an initial state $x_0\in\X_0\subseteq\X$ where it is the robot's turn to act, along with two action sequences
\begin{align*}
\vec{u}_R&=(u_{R,0},u_{R,1},...)\in\U_R^\infty \\
\vec{u}_H&=(u_{H,0},u_{H,1},...)\in\U_H^\infty
\end{align*}
for $R$ and $H$, respectively, the \emph{trajectory} from $x_0$ using $\vec{u}_R,\vec{u}_H$ is the sequence of states 
\begin{align*}
\zeta_R(x_0,\vec{u}_R,\vec{u}_H)=(x_0,x_0',x_1,...)\in\X^\infty,
\end{align*}
where $x_t'=f_R(x_t,u_{R,t})$ and $x_{t+1}=f_H(x_t',u_{H,t})$.
Similarly, given an initial state $x_0'\in\X$ where it is the human's turn to act along with $\vec{u}_R\in\U_R^\infty$ and $\vec{u}_H\in\U_H^\infty$, the trajectory from $x_0$ using $\vec{u}_H,\vec{u}_R$ is the sequence of states 
\begin{align*}
\zeta_H(x_0',\vec{u}_H,\vec{u}_R)=(x_0',x_0,x_1',...)\in\X^\infty,
\end{align*}
where $x_t=f_H(x_t',u_{H,t})$ and $x_{t+1}'=f_R(x_t,u_{R,t})$. Also, given robot and human policies $\pi_R:\X\to\U_R$ and $\pi_H:\X\to\U_H$, respectively, we define the trajectory
\begin{align*}
\zeta_R(x_0,\pi_R,\pi_H)=(x_0,x_0',x_1,...)\in\X^\infty
\end{align*}
by $x_t'=f_R(x_t,\pi_R(x_t))$ and $x_{t+1}=f_H(x_t',\pi_H(x_t'))$, and
\begin{align*}
\zeta_H(x_0,\pi_H,\pi_R)=(x_0',x_0,x_1',...)\in\X^\infty
\end{align*}
by $x_t=f_H(x_t',\pi_H(x_t'))$ and $x_{t+1}'=f_R(x_t,\pi_R(x_t))$.

Given a given safe region $\Xs\subseteq\X$, our goal is to ensure the system stays in $\Xs$ for the entire trajectory.
\begin{definition}
\rm
A trajectory $\zeta=(x_0,x_0',x_1,...)$ (or $\zeta=(x_0',x_0,x_1',...)$) is \emph{safe} if $x_t,x_t'\in\Xs$ for all $t\in\mathbb{N}$.
\end{definition}

\paragraph{Example}

Consider an autonomous driving robot $R$ interacting with a human driver $H$. The state $x\in\X=\mathbb{R}^8$ is a vector $x=(x_R,y_R,v_R,\theta_R,x_H,y_H,v_H,\theta_H)$ representing the positions $(x_R,y_R),(x_H,y_H)$, velocities $v_R,v_H$, and angles $\theta_R,\theta_H$ of $R$ and $H$, respectively. The actions $u\in\U_R=\U_H=\mathbb{R}^2$ are vectors $u=(\phi,a)$ representing the steering angle $\phi$ and acceleration $a$; we assume $|\phi|\le\phi_{\text{max}}$ and $|a|\le a_{\text{max}}$ are bounded. Given time step $\tau\in\mathbb{R}_{>0}$, the dynamics are
\begin{align*}
f_A(x,u_A)&=x+g_A(x,u_A)\cdot\tau\qquad(\forall A\in\{R,H\}) \\
g_R(x,u_R)&=(v_R\cos\theta_R,v_R\sin\theta_R,a_R,v_R\phi_R,0,0,0,0) \\
g_H(x,u_H)&=(0,0,0,0,v_H\cos\theta_H,v_H\sin\theta_H,a_H,v_H\phi_H).
\end{align*}
Safety means the robot and the human have not collided:
\begin{align*}
\Xs=\{x\in\X\mid\|(x_R,y_R)-(x_H,y_H)\|\ge d_{\text{safe}}\},
\end{align*}
for some constant $d_{\text{safe}}\in\mathbb{R}_{\ge0}$.

\section{Safety Modulo Fault}
\label{sec:human}

Ensuring safety in the presence of an adversarial human would be impossible or at least significantly degrade performance---e.g., to avoid an accident with an adversarial human driver, the robot would have to maintain a very large distance. Thus, to ensure safety, we must make assumptions about the behavior of the human. Ideally, we want to make the minimal possible assumptions about the behavior of the human while still accounting for all possible behaviors of a human acting in a responsible way. Then, we want to ensure safety for any human acting according to these assumptions.

The key challenge is devising a reasonable set of assumptions on the human. Intuitively, our assumptions are based on the idea that if the human can act in a safe way to avoid an accident, then they do so (Assumption~\ref{assump:humansafety}). We need to formalize what it means for the human to ``be able to act in a safe way''. We give the human great leeway in what safe actions they consider---for instance, we could assume the human always considers slowing down in \emph{some} manner to ensure safety (Assumption~\ref{assump:humanbackup}). Finally, just as we make assumptions about how the human acts, we expect the human to make assumptions about how the robot may act. We give the human great leeway in doing so---for instance, we could assume that the human always accounts for the possibility that the robot may take a safe action such as gradually braking to avoid an accident (Assumption~\ref{assump:robotbackup}). We formalize our assumptions and safety notion below.

\paragraph{Human objective}

Our definition of safety is based on a model of a human acting according to a maximin objective. In this objective, the ``min'' is the worst-case over a set of action sequences that the human anticipates the robot may take, and the ``max'' is over the human's own actions. That is, the human plans optimally according to their objective, while conservatively accounting for actions they anticipate the robot might take. In particular, suppose that at state $x$, the robot takes action $u_R$ and transitions to state $x'=f_R(x,u_R)$; then, the human takes an action $\pi_H(x')=u_{H,0}^*$ such that
\begin{align}
\label{eqn:human}
\vec{u}_H^*\in\operatorname*{\arg\max}_{\vec{u}_H\in\U_H^\infty}J_H(x',\vec{u}_H),
\end{align}
where the $\arg\max$ denotes the set of all optimal values, and
\begin{align*}
J_H(\vec{u}_H)&=\min_{\vec{u}_R\in\hat{\U}_R^\infty}J_H(\zeta_H(x',\vec{u}_H,\vec{u}_R)) \\
J_H((x_0',x_0,x_1,...))&=\sum_{t=0}^{\infty}\gamma^t r_H(x_t',u_{H,t},x_t),
\end{align*}
where
$\hat{\U}_R\subseteq\mathbb{R}^{n_{U,R}}$ is the set of actions the human anticipates the robot may take,
$r_H:\X\times\U^H\times\X\to\mathbb{R}\cup\{-\infty\}$ is the human reward function,
and $\gamma\in(0,1)$ is a discount factor.

The key challenge for the robot to plan safely is that it does not know the human reward function $r_H$, the human action set $\U_H$ or the human-anticipated robot action set $\hat{\U}_R$. Assuming we know these values exactly is implausible. Instead, we assume access to minimal knowledge about each of these objects. Intuitively, our assumptions are:
\begin{itemize}
\item {\bf Human reward function:} The human always prioritizes avoiding an unsafe state---in particular, their reward function $r_H$ equals $-\infty$ for any unsafe state.
\item {\bf Human-anticipated robot actions:} We are given a conservative robot action $u_R^0$ (e.g., braking) in $\hat{\U}_R$.
\item {\bf Human actions:} We are given a set of human actions $\U_H^0$ such that the human takes \emph{some} action $u_H\in\U_H^0$ whenever they cannot ensure safety.
\end{itemize}
We formalize these assumptions in the next section.

\paragraph{Assumption on human objective}

First, we assume that the human reward for reaching an unsafe state is $-\infty$.
\begin{assumption}
\label{assump:humansafety}
\rm
For any $x',x\in\X$ and $u_H\in\U_H$, we have $r_H(x',u_H,x)=-\infty$ if $x'\not\in\Xs$ or $x\not\in\Xs$.
\end{assumption}
That is, the human driver always acts to avoid an accident. Other than Assumption~\ref{assump:humansafety}, $r_H$ can be arbitrary.

With this assumption, there are two reasons accidents may happen: (i) there was a safe action sequence $\vec{u}_H\in\U_H^\infty$ that the human driver failed to take, or (ii) if the robot takes an action $u_R\not\in\hat{\U}_R$ that the human driver failed to anticipate. Thus, we can always conservatively take $\hat{\U}_R$ to be smaller than it actually is. Conversely, we can always take $\U_H^0$ to be larger than it actually is. Thus, we make minimal assumptions about what actions are contained in $\hat{\U}_R$ and $\U_H$.

First, we make the following assumption on the set of actions $\vec{\U}_R$ that the human anticipates the robot may take:
\begin{assumption}
\label{assump:robotbackup}
\rm
We are given a \emph{robot backup action} $u_R^0\in\U_R$ that is anticipated by the human---i.e., $u_R^0\in\hat{\U}_R$.
\end{assumption}
That is, the human always accounts for the possibility that the robot might take action $u_R^0$. For example, we might assume that $u_R^0$ is gradually braking and coming to a stop.

Next, we make the following assumption about the human:
\begin{assumption}
\label{assump:humanbackup}
\rm
We are given a \emph{human backup action set} $\U_H^0\subseteq\U_H$ such that if $\max_{\vec{u}_H\in\U_H^\infty}J_H(\vec{u}_H)=-\infty$, then the human takes an action $\pi_H(x')\in\U_H^0$.
\end{assumption}
That is, if the human is unable to guarantee safety (i.e., their objective value is $-\infty$), then they take \emph{some} action in $\U_H^0$. For example, $\U_H^0$ may contain all actions where the human driver decelerates by at least some rate (so they can slow down more quickly or steer in any direction).
\begin{remark}
\rm
Our approach can easily be extended to the case where $u_R^0$, $\U_H^0$, and $\hat{\U}_R$ are time varying.
\end{remark}
\begin{remark}
\rm
We can weaken (\ref{eqn:human}) as follows. We only need to assume that the human chooses an action sequence $\vec{u}_H^*$ with a reward $>-\infty$. Then, Assumption~\ref{assump:humansafety} say that the human chooses actions $\vec{u}_H^*$ that will avoid an accident assuming the robot takes actions in $\hat{\U}_R$. For instance, we can replace (\ref{eqn:human}) with a probabilistic model~\cite{ziebart2008maximum} where the human samples action sequence $\vec{u}_H$ with probability $p(\vec{u}_H\mid x_0,u_{R,0})\propto\exp(J_H(\vec{u}_H))$ and takes action $u_{H,0}$; then, Assumption~\ref{assump:humansafety} says the human never considers trajectories that may reach an unsafe state.
\end{remark}

\paragraph{Problem formulation}

Our goal is to ensure that the robot acts in a way that ensures safety for an infinite horizon for any human that satisfies our assumptions.
\begin{definition}
\label{def:safe}
\rm
A robot policy $\pi_R:\X\to\U_R$ is \emph{safe modulo fault} for initial states $\X_0\subseteq\X$ if for any human policy $\pi_H$ satisfying Assumptions~\ref{assump:humansafety}, \ref{assump:robotbackup}, \& \ref{assump:humanbackup}, and any $x_0\in\X_0$, the trajectory $\zeta_R(x_0,\pi_R,\pi_H)\in\X^\infty$ is safe.
\end{definition}
That is, $\pi_R$ that ensures safety as long as the human acts in a way that satisfies our assumptions. Our goal is to design a policy $\pi_R$ that is safe modulo fault. We use the term ``fault'' to indicate that this property only guarantees safety with respect to humans that satisfy our assumptions; we are not assigning blame to the human in case of an accident. The controller designer is responsible for ensuring the assumptions are satisfied by reasonable human drivers.

Finally, we cannot guarantee safety starting from an arbitrary state $x_0$. For instance, if the robot is about to crash into a wall, no action can ensure safety. We assume that the initial states $\X_0$ are ones where we can guarantee safety.
\begin{definition}
\rm
A \emph{safe equilibrium state} $x\in\X$ satisfies (i) $x\in\X_{\text{safe}}$, and (ii) $x=f(x,u_R^0,u_H)$ for all $u_H\in\U_H^0$.
\end{definition}
We denote the set of safe equilibrium states by $\Xe$. At a state $x\in\Xe$, the robot and human can together ensure safety for an infinite horizon by taking actions $u_R^0$ and $u_H$ for any $u_H\in\U_H^0$. In our driving example, $\Xe$ contains states where both agents are at rest (i.e., their velocity is zero).
\begin{assumption}
\rm
We have $\X_0\subseteq\Xe$.
\end{assumption}
In other words, the system starts at a safe equilibrium state where we can ensure safety for an infinite horizon.

\section{Model Predictive Shielding Modulo Fault}
\label{sec:algo}

We describe our algorithm for constructing a robot controller $\pi_R:\X\to\U_R$ that is safe modulo fault. Our approach is based on \emph{shielding}~\cite{alshiekh2018safe}---it takes as input an arbitrary controller $\hat\pi_R:\X\to\U_R$ and modifies it to construct $\pi_R$. Intuitively, $\pi_R$ overrides $\hat\pi_R$ when it cannot ensure it is safe.

\begin{algorithm}[t]
\begin{algorithmic}
\Procedure{$\pi_R$}{$x$}
\State $x'\gets f_R(x,\hat\pi_R(x))$
\State \textbf{return if} $\textsc{IsRec}(x')$ \textbf{then} $\hat\pi_R(x)$ \textbf{else} $u_R^0$ \textbf{end if}
\EndProcedure
\Procedure{IsRec}{$x'$}
\State $X_0'\gets\{x'\}$
\For{$t\in\{0,...,k-1\}$}
\State \textbf{if} $X_t'\not\subseteq\Xs$ \textbf{then return false end if}
\State $U_{R,t}\gets$ {\bf if} $t=0$ {\bf then} $\{u_R\}$ {\bf else} $\{u_R^0\}$ {\bf end if}
\State $U_{H,t}\gets\U_H^0$
\State $X_{t+1}'\gets F(X_t,U_{R,t},U_{H,t})$
\EndFor
\State \textbf{return if} $X_k\subseteq\Xe$ \textbf{then true} \textbf{else false end if}
\EndProcedure
\end{algorithmic}
\caption{Model predictive shielding modulo fault.}
\label{alg:shield}
\end{algorithm}

The challenge is checking whether it is safe to use $\hat\pi_R$. Model predictive shielding (MPS) is an approach to shielding that checks safety online based on the following~\cite{wabersich2018linear,li2019robust}. These approaches are designed to handle deterministic or stochastic environments where the model is known, whereas we do not have any such model of the human driver. We show how to extend these algorithms to our setting. The idea is to maintain the invariant that the current state is \emph{recoverable}---intuitively, that there is some sequence of actions each agent can take that safely brings the system to a stop. In particular:
\begin{definition}
\label{def:recoverable}
\rm
Given hyperparameter $k\in\mathbb{N}$, a state $x'\in\X$ is \emph{recoverable} (denoted $x'\in\Xr$) if for any $\vec{u}_H\in(\U_H^0)^\infty$ and for $\vec{u}_R=(u_R^0,u_R^0,...)\in\U_R^\infty$, the trajectory $\zeta_H(x',\vec{u}_H,\vec{u}_R)=(x_0',x_0,x_1',...)$ satisfies (i) $x_t',x_t\in\Xs$ for all $t\in\{0,...,k\}$, and (ii) $x_k\in\Xe$.
\end{definition}
That is, $x'$ is recoverable if the human and robot can ensure safety by using their respective backup actions $\U_H^0$ and $u_R^0$. By definition, if $x_k\in\Xe$, then $x_k=x_{k+1}'=x_{k+1}=...$. Since $x_k\in\Xs$, it follows that $x_t',x_t\in\Xs$ for all $t\in\mathbb{N}$.

Now, our MPS modulo fault algorithm for computing $\pi_R$ is shown in Algorithm~\ref{alg:shield}. \textsc{IsRec} checks whether $x'=f_R(x,\hat\pi_R(x))$ is recoverable. If so, $\pi_R$ returns $\hat\pi_R(x)$; otherwise, it returns $u_R^0$. The key novelty is how \textsc{IsRec} checks recoverability. Existing approaches~\cite{wabersich2018linear,li2019robust} do so by simulating the model of the environment. In contrast, we need to guarantee safety with respect to \emph{all} $\vec{u}_H\in(\U_H^0)^\infty$. To do so, \textsc{IsRec} \emph{conservatively overapproximates} recoverability---i.e., if it says that $x'$ is recoverable, then it must be recoverable, but it may say $x'$ is not recoverable even if it is.

To do so, \textsc{IsRec} overapproximates the reachable set of states after $t$ steps as a subset $X_t\subseteq\X$. More precisely, it assumes given a \emph{dynamics overapproximation} $F:2^\X\times2^{\U_R}\times2^{\U_H}\to2^\X$ mapping sets of states $X\subseteq\X$, sets of robot actions $U_R\subseteq\U_R$, and sets human action $U_H\subseteq\U_H$ to sets of states $F(X,U_R,U_H)\subseteq2^\X$, which satisfies
\begin{align}
\label{eqn:ai}
f(x,u_R,u_H)\in F(X,U_R,U_H)
\end{align}
for all $x\in X$, $u_R\in U_R$, and $u_H\in U_H$---i.e., $F(X,U_R,U_H)$ contains \emph{at least} the states reachable from $x\in X$ by taking actions $u_R\in U_R$ and $u_H\in U_H$. Intuitively, computing $F$ in a way that this property holds with equality may be computationally intractable, but there exist tractable overapproximations---e.g., based on polytopes~\cite{althoff2010computing,sadraddini2019linear} or ellipsoids~\cite{asselborn2013control,filippova2017ellipsoidal}. This approach fits in the general framework of \emph{abstract interpretation}~\cite{cousot1977abstract}; here, the sets $X$, $U_R$, and $U_H$ are represented implicitly rather than explicitly for computational tractability. We describe the overapproximation we use for our autonomous driving example in Section~\ref{sec:overapproximation}.

Finally, \textsc{IsRec} checks whether (i) safety holds for every state $x_t\in X_t$ (i.e., $X_t\subseteq\Xs$), and (ii) every state $x_k\in X_k$ is a safe equilibrium state (i.e., $X_k\subseteq\Xe$). If both these properties hold, then $x$ is guaranteed to be recoverable. We have the following guarantee (see Appendix~\ref{sec:thmmainproof} for a proof):

\begin{theorem}
\label{thm:main}
Assuming (\ref{eqn:ai}) holds, then our policy $\pi_R$ is safe modulo fault (i.e., it satisfies Definition~\ref{def:safe}).
\end{theorem}

\section{Evaluation}
\label{sec:exp}

We have implemented our approach in a simulation for three robotics tasks. For the robot, we consider an aggressive controller with and without the shield as well as a cross entropy method controller (CEM) that is designed to avoid humans. For the human, we use both simulated humans based on a social forces model of pedestrians~\cite{helbing1995social}, as well as real humans interacting with the simulation via keyboard inputs.

Our goal is to understand how our approach can ensure safety in aggressive driving scenarios. Thus, we focus on settings where the human (simulated or real) and robot must compete to reach their goals. We tune the parameters of our MPS modulo fault algorithm (i.e., the robot backup action $u_R^0$ and the human backup action set $\U_H^0$) to be as aggressive as possible while still ensuring safety on the simulated humans. Furthermore, for our experiments with real-world humans, we strongly encourage them to try and reach their goal before the robot, albeit keeping safety as the top priority. Then, our results are designed to answer the following:
\begin{itemize}
\item Can MPS modulo fault can be used to ensure safety  with real and simulated humans?
\item Can MPS modulo fault outperform a handcrafted MPC based on CEM in terms of performance?
\end{itemize}

\subsection{Experimental Setup}

\paragraph{Robotics tasks}

\begin{figure}
\centering
\begin{tabular}{cc}
\includegraphics[height=0.18\textwidth]{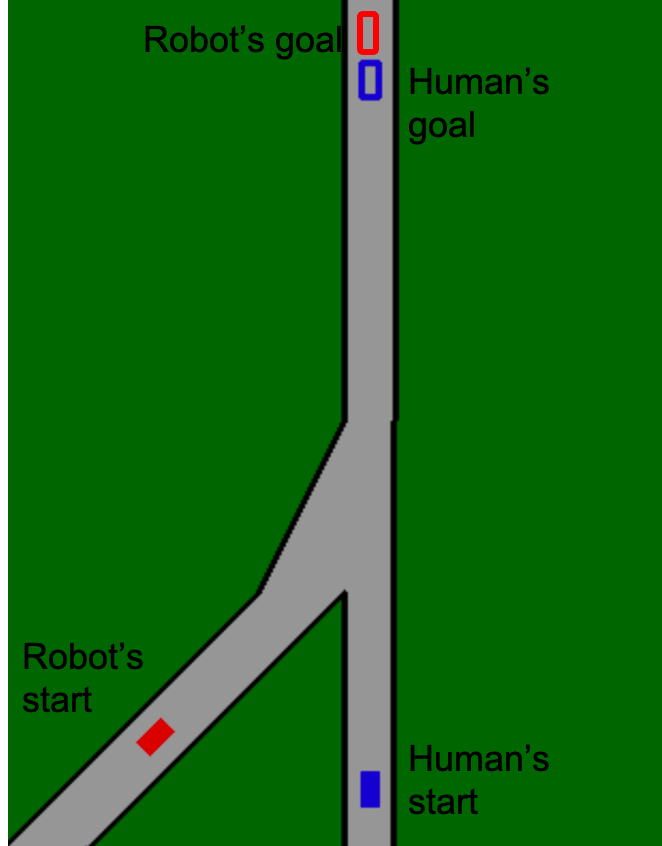} &  
\includegraphics[height=0.18\textwidth]{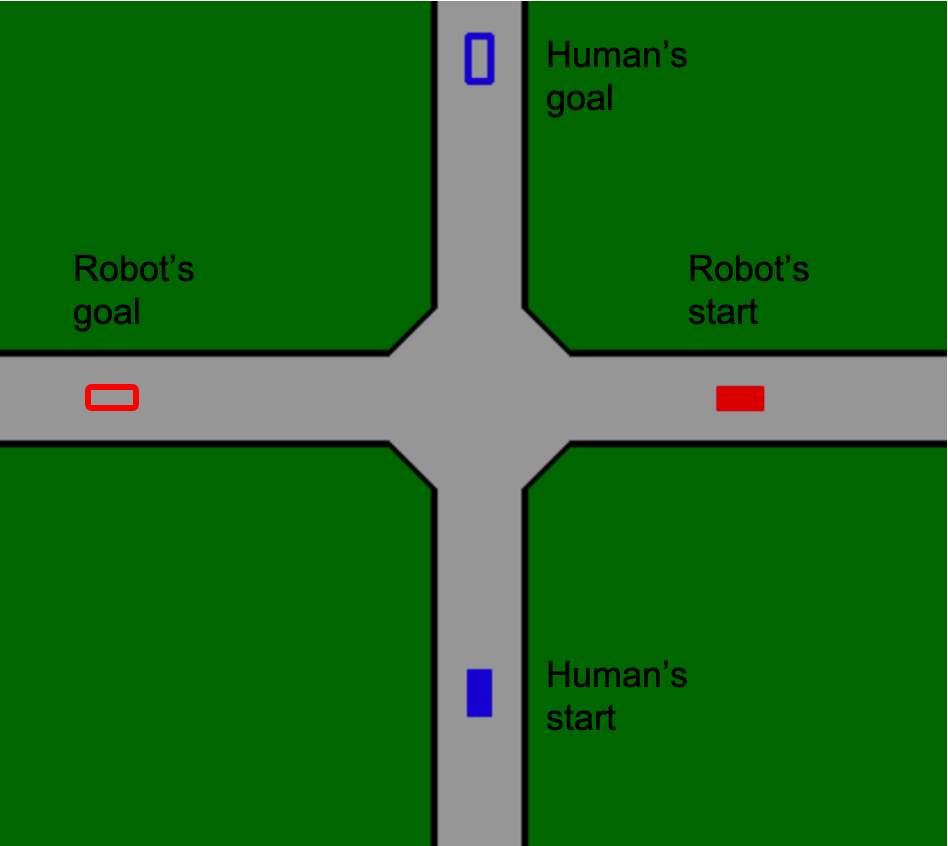}  \\
(a) merge  & (b) cross \\
\end{tabular}
\begin{tabular}{ccc}
\includegraphics[height=0.18\textwidth]{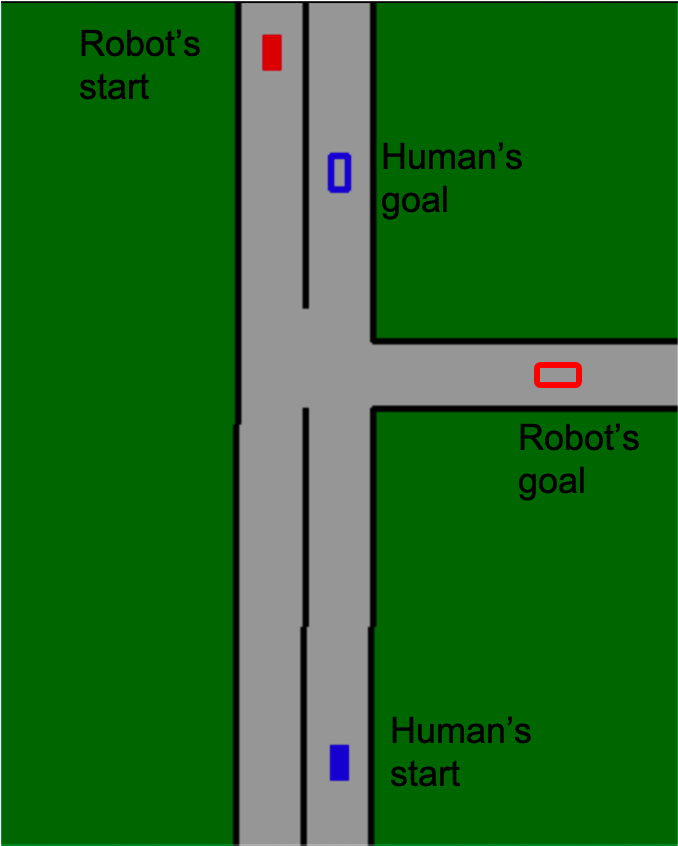} &  
\includegraphics[height=0.18\textwidth]{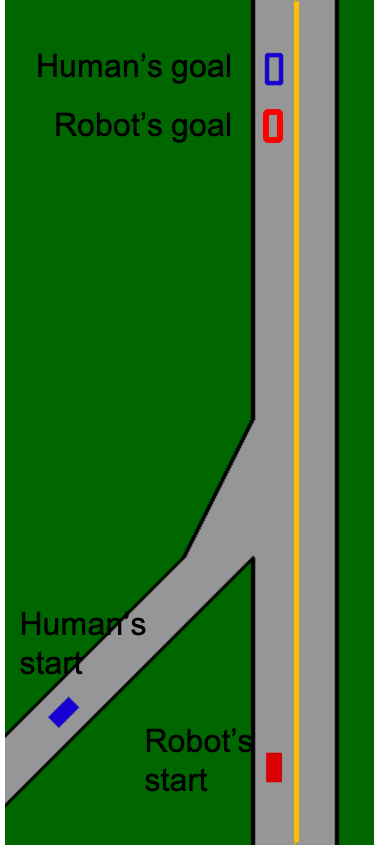} &  
\includegraphics[height=0.18\textwidth]{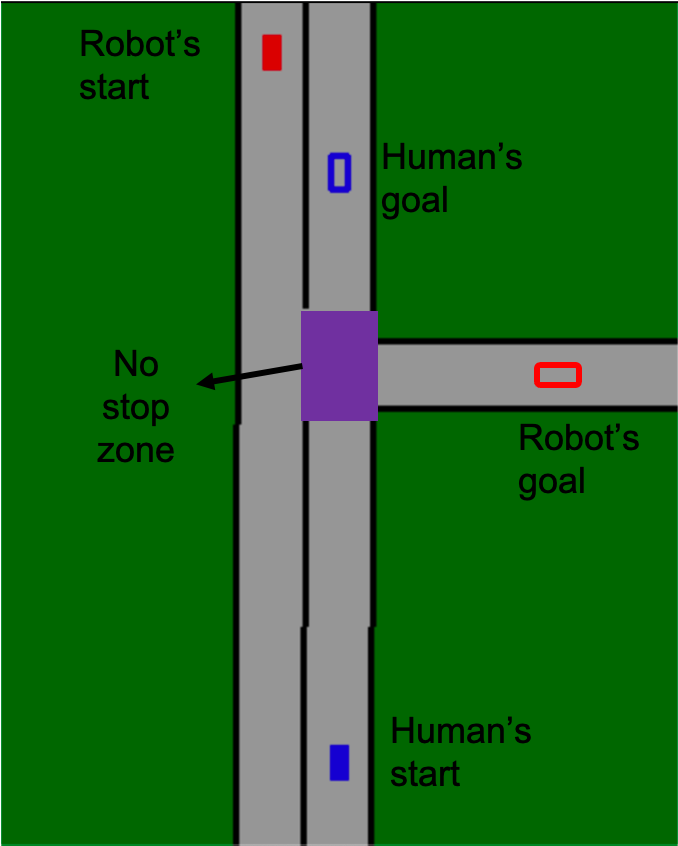} \\
(c) turn & (d)  two lanes & (e) turn (no stop) 
\end{tabular}
\caption{Visualizations of tasks along with the initial positions and goals of the robot and human. The red (resp., blue) box is the robot (resp., human).}
\label{fig:tasks}
\end{figure}

We consider three non-cooperative robotics tasks (depicted in Figure~\ref{fig:tasks}). In the first task (``merge''), there are two lanes that merge---i.e., the robot is coming in from one lane and the humans from another; the robot and human goals are to navigate the merge and reach their goal. The second task (``cross'') has both the human and the robot moving towards an intersection from different directions---i.e., the robot is moving horizontally and the human is moving vertically; the robot and human goals are to get to their goal on the other side of the intersection. The third task (``turn'') is an unprotected left turn---i.e., the humans are driving without turning and the  robot needs to make a left turn that crosses the human path.

\paragraph{Safety property}

We assume the robot and human are each a rectangle; then, the safety property is that the the robot and human rectangles should not intersect.

\paragraph{Robot dynamics}

The robot dynamics are the ones in our running example---i.e., its state is $(x,y,v,\theta)$, where $(x,y)$ is position, $v$ is velocity, and $\theta$ is orientation, and its actions are $(a,\phi)$, where $a$ is acceleration and $\phi$ is steering angle. We assume $|a|\le a_{\text{max}}$, $|\phi|\le\phi_{\text{max}}$, and $0\le v\le v_{\text{max}}$.

\paragraph{Simulated humans}

For simulated humans, we use the social force model~\cite{helbing1995social}, which includes potential forces that cause each human to avoid the robot, other humans, and walls, while trying to reach their goal.

\paragraph{Real humans}

We also considered real human users interacting with the simulation via keyboard. They control the human using the up/down arrows to control acceleration and the left/right arrows to control steering angle. We asked the human users to prioritize safety first, but to drive aggressively to try and reach their goal before the robot.

\paragraph{Controllers}

We consider three controllers for the robot: (i) an aggressive controller, (ii) a handcrafted MPC based on the cross-entropy method (CEM) designed to ensure safety without a shield, and (iii) our MPS modulo fault algorithm used in conjunction with the aggressive controller.

The first controller is an ``aggressive controller'' that ignores the humans and moves directly towards the goal as quickly as possible. If nonlinear trajectories are required, we manually specify a sequence of subgoals; once the robot reaches its current subgoal, it continues to the next one.

The second is a model-predictive controller (MPC) that aims to avoid colliding with the human. We use a planning algorithm based on the cross-entropy method (CEM). Then, it chooses the action that attempts to optimize its objective over the planning horizon. We use a handcrafted objective that provides a positive reward for progressing towards its goal and a large negative penalty for collisions. To predict collisions, it forecasts the behavior of the human over the planning horizon by extrapolating their position based on their current velocity (i.e., constant velocity assumption). Finally, for the goal-reaching portion of the objective, we use subgoals the same way we do for the aggressive controller.

The third controller is our MPS modulo fault algorithm used with the aggressive controller. The robot backup action is $u_R^0  =  (0, -1)$ where $\phi=0$ is the steering angle and $a=-1$ is the acceleration. The human backup action set is
\begin{align*}
\U_H^0  = \left\{(\phi,a) \biggm\vert \phi \in \left[  - \frac{\pi}{10}, \frac{\pi}{10} \right],~ a \in \left[-1, -\frac{1}{2}\right] \right\}.
\end{align*}
That is, the human predicts that the robot may gradually brake without changing direction, and the human considers braking gradually (or hard) while steering up to some angle.

\subsection{Experimental Results}

We describe our experimental results. For simulated humans, all results shown are averaged over 100 simulations. For real humans, the results are based on 18 users.

\paragraph{MPS modulo fault ensures safety for simulated humans}

\begin{figure}
\centering
\includegraphics[width=0.23\textwidth]{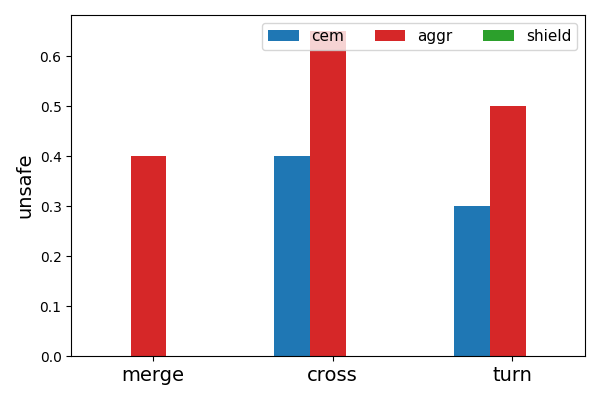}
\includegraphics[width=0.23\textwidth]{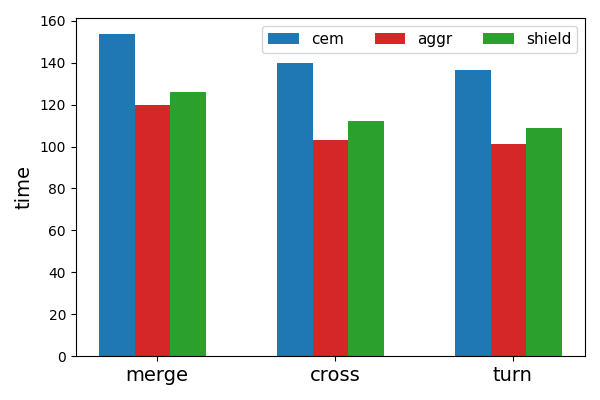}
\caption{Results with simulated humans, for the aggressive controller (red), the CEM MPC (blue), and our shielded controller (green). Left: Fraction of unsafe runs. Right: Time the robot takes to reach its goal in seconds.}
\label{fig:sim_results}
\end{figure}

In Figure~\ref{fig:sim_results}, we show both the fraction of unsafe runs (left), and the time  taken by the robot to  reach the goal (right), including the aggressive controller (red), the MPC based on CEM (blue), and our shielded aggressive controller (green). As can be seen, for the aggressive controller, the rate of unsafe runs is very high since the robot ignores the human to get to its goal. Next, for  the MPC based on CEM, the rate of unsafety is lower but still not zero. However, the CEM policy takes significantly longer to reach its goal compared to the aggressive policy. Finally, our shielded aggressive controller is always safe, yet only takes a small amount of time longer to reach its goal compared to the aggressive policy; in particular, it is significantly faster than the MPC. These comparisons demonstrate that our approach greatly improves safety without significantly reducing time to goal.

\paragraph{MPS modulo fault ensures safety for real humans}

\begin{figure}
\centering
\includegraphics[width=0.23\textwidth]{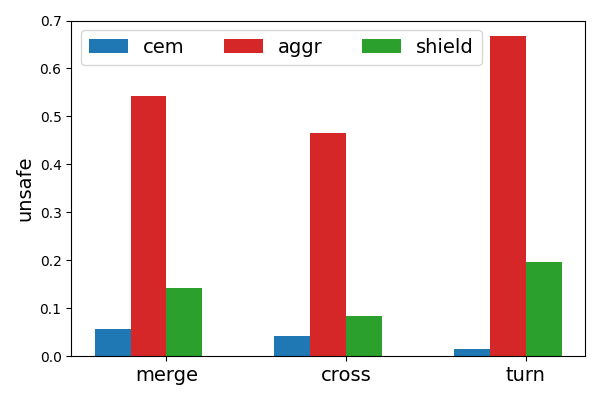}
\includegraphics[width=0.23\textwidth]{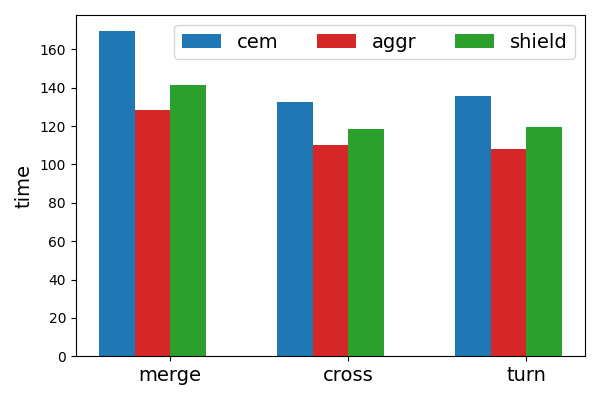}
\caption{Results with real humans, for the aggressive controller (red), the CEM MPC (blue), and our shielded controller (green). Left: Fraction of unsafe runs. Right: Time the robot takes to reach its goal in seconds.}
\label{fig:human_results}
\end{figure}

Next, we had real human users interact with our simulated robot via keyboard input. we show both the fraction of unsafe runs (left), and the time taken by the robot to reach the goal (right), including the aggressive controller (red), the MPC based on CEM (blue), and our shielded aggressive controller (green). As can be seen, for the aggressive controller, the robot gets to its goal the fastest, but is frequently unsafe. The MPC based on CEM is significantly safer; in this case, it is somewhat safer than our shielded aggressive controller. On the other hand, our shield controller reaches its goal significantly faster than the MPC. As described above, we set the shield parameters aggressively based on the simulated humans to ensure it could reach its goal; in practice, we could ensure safety by setting these parameters more conservatively and by tuning them to the real human driver data.

\begin{figure}
\centering
\includegraphics[width=0.1\textwidth]{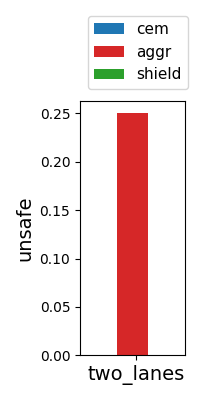} 
\includegraphics[width=0.1\textwidth]{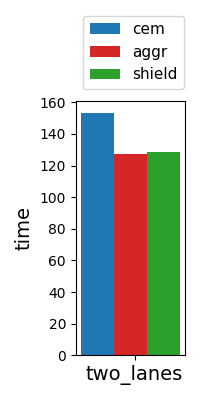} 
\includegraphics[width=0.1\textwidth]{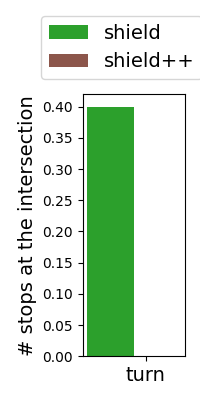} 
\includegraphics[width=0.1\textwidth]{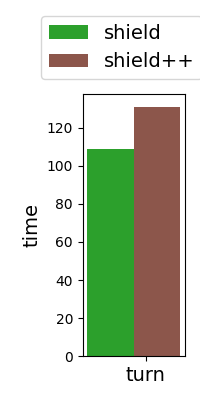} 
\caption{Results for alternative robot backup actions with simulated humans. For the ``pull over'' backup action, we show the fraction of unsafe runs (leftmost) and the time the robot takes to reach its goal in seconds (second from the left), for the aggressive controller (red), the CEM MPC (blue), and our shielded aggressive controller (green). For the ``no-stop zone'' backup action, we show the number of stops in the intersection (second from the right) and the time the robot takes to reach its goal in seconds (rightmost), for the original (green, ``shield'') and the new (brown, ``shield++'') shielded controllers; both controllers are always safe.}
\label{fig:extra_results}
\end{figure}

\paragraph{Alternative robot backup actions}

A key feature of our approach is that we can flexibly design the robot backup action to ensure safety. To demonstrate this flexibility, we design an alternative backup action that pulls the robot over to the shoulder of a highway. This backup policy is time varying---i.e., the robot steering depends on the current state. We test this backup policy with simulated humans on the task in Figure~\ref{fig:tasks} (d), where there are two lanes on the highway and an on-ramp that merges onto the highway. The human is on the on-ramp and the robot is on  the highway. To avoid collisions, the robot can  pull over to the right-most lane. Figure~\ref{fig:extra_results} shows  the  fraction of the unsafe runs (leftmost), and the time the robot takes to reach its goal (second from left) for all three controllers---aggressive (red), the MPC based on CEM (blue), and our shielded aggressive controller with the pull over backup policy (green). Our shielded controller is always safe and is   significantly faster than the MPC.

We also design a robot backup action that avoids stopping in the middle of an intersection and blocking it, which is often illegal. To this end, we modify the turn task to include a no-stop zone (shown in Figure~\ref{fig:tasks} (e)) where the robot is prohibited from stopping. In this zone, the robot backup action does not come to a stop immediately; instead, it drives through the zone and only brakes once it has fully cleared the intersection. The results for this experiment using simulated humans are shown in Figure~\ref{fig:extra_results} (right). We compare the original shielded controller (``shield'') that may stop in the intersection with the new one that adheres to the no-stop zone in Figure~\ref{fig:tasks} (e) (``shield++''). In  this case, both the controllers were always safe; instead, we show the fraction of runs where the robot stops in the intersection (second from the right), and the time the robot takes to reach its goal (rightmost). The new shielded controller takes slightly longer to reach the goal, but never stops in the intersection.

\section{Conclusion}

We have proposed an approach for ensuring safety in human-interactive robotics systems. We define a notion of safety that models human behavior by specifying their backup behaviors, and propose our MPS modulo fault algorithm for ensuring safety with respect to this model. We have validate our approach on both real and simulated humans.

\bibliographystyle{IEEEtran}
\bibliography{main}

\clearpage
\appendices
\section{Overapproximating the Dynamics}
\label{sec:overapproximation}

We describe our approach for overapproximating the multi-agent dynamics $f$ in our autonomous car example. In particular, we describe representations $U_R$, $U_H$, and $X$ of sets of robot actions, human actions, and states, respectively, along with a function $F:(U_R,U_H,X)\mapsto X'$ that satisfies (\ref{eqn:ai}). First, we consider a representation of $U_R$ of the form
\begin{align}
\label{eqn:urai}
U_R=\{(-a_R,0)\},
\end{align}
where $a_R\in\mathbb{R}_{>0}$ is the rate at which the robot decelerates; we assume it corresponds to the robot gradually braking to give a trailing human driver sufficient time to respond. Also, we consider a representation of $U_H$ of the form
\begin{align}
\label{eqn:uhai}
U_H=(-\infty,-a_{\text{min}}]\times[-\phi_{\text{max}},\phi_{\text{max}}]
\end{align}
where $a_{\text{min}}\in\mathbb{R}_{>0}$ is the minimum deceleration rate---i.e., the minimum rate at which we expect the human driver to brake to avoid an accident. Then, both $U_R$ and $U_H$ have form
\begin{align*}
U_A=[a_{A,\text{min}},a_{A,\text{max}}]\times[-\phi_{\text{max}},\phi_{\text{max}}]
\end{align*}
for $a_{A,\text{min}},a_{A,\text{max}}\in\mathbb{R}$ and $\phi_{\text{max}}\in\mathbb{R}_{>0}$, and where $A\in\{R,H\}$. Next, we consider $X$ of the form
\begin{align}
\label{eqn:xai}
X&=X_R\times X_H \\
X_A&=[x_{A,\text{min}},x_{A,\text{max}}]\times[y_{A,\text{min}},y_{A,\text{max}}] \nonumber \\
&\qquad\times[v_{A,\text{min}},v_{A,\text{max}}]\times[-\theta_{A,\text{max}},\theta_{A,\text{max}}], \nonumber
\end{align}
where $A\in\{R,H\}$. Then, we use
\begin{align*}
F(X,U_R,U_H) &= F_R(X,U_R,U_H) \times F_H(X,U_R,U_H) \\
F_A(X,U_R,U_H) &= X_A^x \times X_A^y \times X_A^v \times X_A^\theta,
\end{align*}
where
\begin{align*}
X_A^x &= [x_{A,\text{min}}-v_{A,\text{max}},x_{A,\text{max}}+v_{A,\text{max}}] \\
X_A^y &= [y_{A,\text{min}}-v_{A,\text{max}},y_{A,\text{max}}+v_{A,\text{max}}] \\
X_A^v &= [\max\{v_{A,\text{min}}+a_{A,\text{min}},0\},\max\{v_{A,\text{max}}+a_{A,\text{max}},0\}] \\
X_A^\theta &= [-\theta_{A,\text{max}}-v_{A,\text{max}}\phi_{A,\text{max}},\theta_{A,\text{max}}+v_{A,\text{max}}\phi_{A,\text{max}}]
\end{align*}
where again $A\in\{R,H\}$. It is easy to see that this choice of $F$ satisfies (\ref{eqn:ai}). Finally, we note that (\ref{eqn:xai}), (\ref{eqn:urai}), and (\ref{eqn:uhai}) are satisfied for all time steps. In particular, $X_0$ satisfies (\ref{eqn:xai}), $U_{R,t}$ satisfies (\ref{eqn:urai}) for all time steps $t$, and $U_{H,t}$ satisfies (\ref{eqn:uhai}) for all time steps $t$ according to our choices in Section~\ref{sec:exp}. Thus, $X_{t+1}=F(X_t,U_R,U_H)$ satisfies (\ref{eqn:xai}) by induction.

\begin{remark}
\rm
Formally, the abstract domain of $U_H$ is $a\in\mathbb{R}\cup\{-\infty,\infty\}$ (with the lattice structure $a\sqsubseteq a'$ if and only if $a\le a'$, $\top=\infty$, and $\bot=-\infty$), and its abstraction and concretization functions are $\alpha:U_H\mapsto\sup_{a\in\mathbb{R}}U_H$ and $\gamma:a\mapsto(-\infty,-a]$, respectively. The setup for $U_R$ is similar. Then, (\ref{eqn:ai}) says that $F$ is an abstract transformer for $f$.
\end{remark}

\section{Proof of Theorem~\ref{thm:main}}
\label{sec:thmmainproof}

First, we have the following guarantee on $F$.
\begin{lemma}
\label{lem:abstract}
Given initial state $x_0\in\X$, and robot and human action set sequences $(U_{R,0},U_{R,1},...)\in(2^{\U_R})^\infty$ and $(U_{H,0},U_{H,1},...)\in(2^{\U_H})^\infty$, respectively, let the state set sequence $(X_0,X_1,...)\in(2^\X)^\infty$ be defined by $X_0=\{x_0\}$ and $X_{t+1}=F(X_t,U_{R,t},U_{H,t})$ for $t\in\mathbb{N}$. Then, for any $\vec{u}_R\in\U_R^{\infty}$ and $\vec{u}_H\in\U_H^{\infty}$ such that $u_{R,t}\in U_{R,t}$ and $u_{H,t}\in U_{H,t}$ for all $t\in\mathbb{N}$, the trajectory $(x_0,x_1,...)\in\X^\infty$ from $x_0$ using $\vec{u}_R,\vec{u}_H$ satisfies $x_t\in X_t$ for all $t\in\mathbb{N}$.
\end{lemma}

\textit{Proof:} We prove by induction on $t$. The base case $t=0$ follows since $x_0=x\in X_0$. For the inductive case, note that
\begin{align*}
x_{t+1}&=f(x_t,u_{R,t},u_{H,t})\in F(X_t,U_{R,t},U_{H,t})=X_{t+1},
\end{align*}
since $x_t\in X_t$ (by induction), since $u_{R,t}\in U_{R,t}$ and $u_{H,t}\in U_{H,t}$ (by assumption), and by (\ref{eqn:ai}). $\qed$

Next, we have the following guarantee on \textsc{IsRec}.
\begin{lemma}
\label{lem:recoverable}
If $\textsc{IsRec}(x')=\text{true}$, then $(x')\in\Xr$.
\end{lemma}

\textit{Proof:}
Suppose $\textsc{IsRec}(x')$ returns true. We need to show that for any $\vec{u}_H=\U_H^\infty$ and for $\vec{u}_R=(u_R^0,u_R^0,...)\in\U_R^\infty$, the trajectory $\zeta_H(x',\vec{u}_H,\vec{u}R)=(x_0',x_0,x_1',...)\in\X^\infty$ satisfies (i) $x_t\in\Xs$ for all $t\in\{0,...,k\}$, and (ii) $x_k\in\Xe$. For (i), $x_t\in X_t\subseteq\Xs$, where the first inclusion follows by Lemma~\ref{lem:abstract} and the second follows by the definition of \textsc{IsRec}. For (ii), as in the previous case, $x_k\in X_k\subseteq\Xe$. $\qed$

Next, we prove the existence of a safe invariant set---i.e., one preserved by $\pi_R$ and any $\pi_H$ satisfying our assumptions.
\begin{definition}
\rm
A state $x'\in\X$ where it is the human's turn to act is \emph{invariant} (denoted $x'\in\Xinv$) if either (i) $J_H(x')>-\infty$, or (ii) $x_+'\in\Xr$, where we have $x=f_H(x',\pi_H(x'))$ and $x_+'=f_R(x,\pi_R(x))$.
\end{definition}
That is, either the human currently has objective value $>-\infty$, or on their next step, the human will be in a recoverable state. Next, we show that $\Xinv$ is an invariant set. We begin with three
technical lemmas. First, we show that if the current human state $x'$ (i.e., it is the human's turn to act) is recoverable and the human uses one of their backup actions $u_H\in\U_H^0$, then the next human state $x_+'$ is also recoverable.
\begin{lemma}
\label{lem:helper1}
If $x'\in\Xr$ and $u_H\in\U_H^0$, then $x\in\Xs$ and $x_+'\in\Xr$, where $x=f_H(x',u_H)$ and $x_+'=f_R(x,\pi_R(x))$.
\end{lemma}
\textit{Proof:} To show $x_+'\in\Xr$, we need to show that for any
\begin{align*}
\vec{u}_H&=(u_{H,0},u_{H,1},...)\in(\U_H^0)^\infty \\
\vec{u}_R&=(u_R^0,u_R^0,...)\in\U_R^\infty,
\end{align*}
the trajectory $\zeta_H(x_+',\vec{u}_H,\vec{u}_R)=(x_0',x_0,x_1',...)$ satisfies (i) $x_t',x_t\in\Xs$ for $t\in\{0,...,k\}$, and (ii) $x_k\in\Xe$. Let 
\begin{align*}
\vec{\tilde{u}}_H&=(u_H,u_{H,0},u_{H,1},...)\in(\U_H^0)^\infty \\
\vec{\tilde{u}}_R&=(u_R^0,u_R^0,...)\in\U_R^\infty,
\end{align*}
where the first line follows since by assumption, we have $u_H\in\U_H^0$. Also by assumption, we have $x'\in\Xr$, so by definition or recoverability, the trajectory $\zeta_H(x',\vec{\tilde{u}}_H,\vec{\tilde{u}}_R)=(\tilde{x}_0',\tilde{x}_0,\tilde{x}_1',...)$ satisfies $\tilde{x}_t',\tilde{x}_t\in\Xs$ for $t\in\{0,...,k\}$ and $\tilde{x}_k\in\Xe$. By definition of equilibrium, $\tilde{x}_{k+1}'=f_H(\tilde{x}_k,\tilde{u}_{H,k})=\tilde{x}_k$ (since $\tilde{u}_{H,k}\in\U_H^0$) and $\tilde{x}_{k+1}=f_R(\tilde{x}_{k+1}',\tilde{u}_{R,k+1})=\tilde{x}_{k+1}'=\tilde{x}_k$. Thus, we have $\tilde{x}_{k+1}',\tilde{x}_{k+1}\in\Xs$ and $\tilde{x}_{k+1}\in\Xe$. Finally, note that $x_t'=\tilde{x}_{t+1}'$ and $x_t=\tilde{x}_{t+1}$. By this fact and the above, both conditions (i) and (ii) holds, so it follows that $x_+'\in\Xr$. Finally, to see $x\in\Xs$, note that $x=\tilde{x}_0\in\Xs$. $\qed$

Next, we show that if our algorithm uses $\hat\pi$ at state $x$, then the next human state $x'$ is recoverable
\begin{lemma}
\label{lem:helper2}
If $\pi_R(x)\neq u_R^0$, then $f_R(x,\pi_R(x))\in\Xr$.
\end{lemma}
\textit{Proof:} By definition of $\pi_R$, if $\pi_R(x)\neq u_R^0$, then we have $\textsc{IsRec}(f_R(x,\hat\pi_R(x)))=\text{true}$. Thus, by Lemma~\ref{lem:recoverable}, $f_R(x,\hat\pi_R(x))\in\Xr$, as claimed. $\qed$

Third, letting $J_H(x')=\max_{\vec{u}_H\in\U_H^\infty}J_H(x',\vec{u}_H)$ be the the human objective value at human state $x'$, we show that if $J_H(x')>-\infty$ and the robot uses its backup action $u_R^0$, then we also have $J_H(x_+')>-\infty$ at the next human state $x_+'$.
\begin{lemma}
\label{lem:helper3}
If $J_H(x')>-\infty$, then $J(x_+')>-\infty$, where $x=f_H(x',\pi_H(x'))$ and $x_+'=f_R(x,u_R^0)$.
\end{lemma}
\textit{Proof:} If $J_H(x')>-\infty$, then $J_H(x',\vec{u}_H^*)>-\infty$, where $\vec{u}_H^*\in\operatorname{\arg\max}_{\vec{u}_H\in\U_H^\infty}J_H(x',\vec{u}_H)$ is the optimal action sequence chosen by $\pi_H$---i.e., $\pi_H(x')=u_{H,0}^*$. Then, we have
\begin{align}
-\infty
&<J_H(x',\vec{u}_H^*) \nonumber \\
&=\min_{\vec{u}_R\in\hat\U_R^\infty}J_H(\zeta_H(x',\vec{u}_H^*,\vec{u}_R) \nonumber \\
&\le\min_{\vec{u}_R\in\{u_R^0\}\times\hat\U_R^\infty}J_H(\zeta_H(x',\vec{u}_H^*,\vec{u}_R)) \nonumber \\
&=r_H(x',u_{H,0}^*,x)+\gamma\min_{\vec{u}_R'\in\hat\U_R^\infty}J_H(\zeta_H(x_+',\vec{\tilde{u}}_H,\vec{u}_R')) \nonumber \\
&=r_H(x',u_{H,0}^*,x)+\gamma J_H(x_+',\vec{\tilde{u}}_H), \label{eqn:lem:helper3:1}
\end{align}
where $\vec{\tilde{u}}_H=(u_{H,1}^*,u_{H,2}^*,...)$. The first line holds by assumption, and the second and last lines hold by definition. The third line follows since by Assumption~\ref{assump:robotbackup}, we have $u_R^0\in\hat\U_R^0$, so $\{u_R^0\}\times\hat\U_R^\infty\subseteq\hat\U_R^\infty$. To see why the fourth line follows, note that for any $\vec{u}_R\in\{u_R^0\}\times\hat\U_R^\infty$, we have\footnote{Here, $\vec{z}\circ\vec{z}'$ denotes the concatenation of sequences $\vec{z}$ and $\vec{z}'$.}
\begin{align*}
\zeta_H(x',\vec{u}_H,\vec{u}_R)
&=(x_0',x_0)\circ(x_1',x_1,x_2',...) \\
&=(x_0',x_0)\circ\zeta_H(x_+',\vec{\tilde{u}}_H,\vec{\tilde{u}}_R),
\end{align*}
where the second line follows since $\vec{u}_R\in\{u_R^0\}\times\hat\U_R^\infty$ implies that $u_{R,0}=u_R^0$, so $x_0=f_H(x_0',u_{H,0}^*)=f_H(x',\pi_H(x'))=x$ and $x_1'=f_R(x_0,u_R^0)=f_R(x,u_R^0)=x_+'$, and since we have $\vec{u}_H=(u_{H,0}^*)\circ\vec{\tilde{u}}_H^*$ and $\vec{u}_R=(u_R^0)\circ\vec{\tilde{u}}_R$, where $\vec{\tilde{u}}_R=(u_{R,1},u_{R,2},...)$. As a consequence, we have
\begin{align*}
&J_H(\zeta_H(x',\vec{u}_H,\vec{u}_R)) \\
&=\sum_{t=0}^\infty\gamma^tr_H(x_t',u_{H,0}^*,x_t) \\
&=r_H(x',u_{H,0}^*,x)+\gamma\sum_{t=0}^\infty\gamma^tr_H(x_{t+1}',u_{H,t+1}^*,x_{t+1}) \\
&=r_H(x',u_{H,0}^*,x)+\gamma J_H(\zeta_H(x_+',\vec{\tilde{u}}_H,\vec{\tilde{u}}_R).
\end{align*}
Then, the fourth line above holds since for any $\vec{\tilde{u}}_R\in\hat\U_R^0$, we have $(u_R^0)\circ\vec{\tilde{u}}_R\in\{u_R^0\}\times\hat\U_R^0$, so 
\begin{align*}
&\min_{\vec{u}_R\in\{u_R^0\}\times\hat\U_R^\infty}J_H(\zeta_H(x',\vec{u}_H,\vec{u}_R)) \\
&=r_H(x',u_{H,0}^*,x)+\gamma\min_{\tilde{\vec{u}}_R\in\hat\U_R^\infty}J_H(\zeta_H(x_+',\vec{\tilde{u}}_H,\vec{\tilde{u}}_R).
\end{align*}
Now, we prove the lemma. Since $r_H(x',u_{H,0}^*,x)<\infty$ and $\gamma>0$, by (\ref{eqn:lem:helper3:1}), we have $J_H(x_+',\vec{\tilde{u}}_H)>-\infty$. Thus, we have
\begin{align*}
J_H(x_+')=\max_{\vec{u}_H'\in\U_H^\infty}J_H(x_+',\vec{u}_H')\ge J_H(x_+',\vec{\tilde{u}}_H)>-\infty,
\end{align*}
as claimed. $\qed$

Next, we show that $\Xinv$ is an invariant set.
\begin{lemma}
\label{lem:invariant}
Suppose $\pi_R$ is as in Algorithm~\ref{alg:shield} and $\pi_H$ satisfies Assumptions~\ref{assump:humansafety}, \ref{assump:robotbackup}, \&~\ref{assump:humanbackup}. If $x'\in\Xinv$, then $x_+'\in\Xinv$ and $x_+',x_+\in\Xs$, where $x=f_H(x',u_H)$, $x_+'=f_R(x,\pi_R(x))$, and $x_+=f_H(x_+',\pi_H(x_+'))$.
\end{lemma}
\textit{Proof:} Since $x'\in\Xinv$, either $x_+'\in\Xr$ or $J(x')>-\infty$. First, suppose that $x_+'\in\Xr$. Let $x_{++}'=f_R(x_+,\pi_R(x_+))$. If $\pi_H(x_+')\in\U_H^0$, then by Lemma~\ref{lem:helper1}, we have $x_{++}'\in\Xr$, so $x_+'\in\Xinv$. Also, in this case, $x_+'\in\Xr\subseteq\Xs$, and by Lemma~\ref{lem:helper1}, $x_+\in\Xs$. Otherwise, if $\pi_H(x_+')\not\in\U_H^0$, then by Assumption~\ref{assump:humanbackup}, we have $J_H(x_+')>-\infty$, so $x_+'\in\Xinv$.
Since $J_H(x_+')>-\infty$, we also have $r_H(x_+',\pi_H(x_+'),x_+)>-\infty$, so by Assumption~\ref{assump:humansafety}, we have $x_+',x_+\in\Xs$. Thus, the claim follows for the case $x_+'\in\Xr$.

Next, suppose that $J(x')>-\infty$. Then, we must have $\pi_R(x)=u_R^0$---otherwise, by Lemma~\ref{lem:helper2}, we would have $x_+'=f_R(x,\pi_R(x))\in\Xr$. Thus, by Lemma~\ref{lem:helper3}, we have $J(x_+') > -\infty$, so $x_+'\in\Xinv$. As before, $J_H(x_+')>-\infty$ also implies that $x_+',x_+\in\Xs$. Thus, the claim follows. $\qed$

Finally, we prove Theorem~\ref{thm:main}.

\textit{Proof:} We can equivalently consider the setting where the human starts at state $x_0'=x$ and takes any action $\pi_H(x_0')\in\U_H^0$; in particular, since $x\in\Xe$, we have $x=f_H(x_0',\pi_H(x_0'))=x_0'$. Now, let $\zeta_H(x_0',\pi_H,\pi_R)=(x_0',x_0,x_1',...)$, where $x_0=x$. Then, note that $x_1'\in\Xr$, since either $\pi_R(x)\neq u_R^0$, in which case $x_1'=f_R(x_0,\pi_R(x_0))\in\Xr$ by Lemma~\ref{lem:helper2}, or $\pi_R(x)=u_R^0$, in which case $x_1'=f_R(x_0,u_R^0)=x_0\in\Xe\subseteq\Xr$. As a consequence, by definition, we have $x_0'\in\Xinv$.

Now, by Lemma~\ref{lem:invariant}, we have $x_t'\in\Xinv$ and $x_{t+1}',x_{t+1}\in\Xs$ for all $t\in\mathbb{N}$. By definition, $x_0'=x_0\in\Xe\subseteq\Xs$. Thus, $x_t',x_t\in\Xs$ for all $t\in\mathbb{N}$, as claimed. $\qed$


\end{document}